\documentclass[10pt,twocolumn,letterpaper]{article}

\usepackage{times}
\usepackage{epsfig}
\usepackage{graphicx}
\usepackage{amsmath}
\usepackage{amssymb}
\usepackage{authblk}

\usepackage[breaklinks=true,bookmarks=false]{hyperref}

\def\iccvPaperID{8718} 

\begin{document}


\title{Unsupervised domain adaptation via coarse-to-fine feature alignment method using contrastive learning}

\author[1]{Shiyu Tang \thanks{Corresponding author: shiyu00daisy@gmail.com}}
\author[1]{Peijun Tang}
\author[1]{Yanxiang Gong}
\author[1]{Zheng Ma}
\author[1]{Mei Xie \thanks{Corresponding author: mxie@uestc.edu.cn
}}
\affil[1]{Department of Computer Science, University of Electronic Science and Technology of China}


\renewcommand*{\Affilfont}{\small\it} 
\renewcommand\Authands{ and } 
\date{} 

\maketitle

\maketitle

\begin{abstract}
Previous feature alignment methods in Unsupervised domain adaptation(UDA) mostly only align global features without considering the mismatch between class-wise features. In this work, we propose a new coarse-to-fine feature alignment method using contrastive learning called CFContra. It draws class-wise features closer than coarse feature alignment or class-wise feature alignment only, therefore improves the model's performance to a great extent. We build it upon one of the most effective methods of UDA called entropy minimization~\cite{Vu_2019_CVPR} to further improve performance. In particular, to prevent excessive memory occupation when applying contrastive loss in semantic segmentation, we devise a new way to build and update the memory bank. In this way, we make the algorithm more efficient and viable with limited memory. Extensive experiments show the effectiveness of our method and model trained on the GTA5~\cite{Richter_2016_ECCV} to Cityscapes dataset has boost mIOU by 3.5 compared to the MinEnt algorithm~\cite{Vu_2019_CVPR}. Our code will be publicly available.

\end{abstract}

\section{Introduction}
Nowadays, semantic segmentation based on deep learning models has been a great success through deeper models like deeplabv3~\cite{DBLP:journals/corr/ChenPSA17} and large datasets like PASCAL VOC~\cite{pascal-voc-2012}, COCO~\cite{lin2014microsoft}, so forth.
However, it takes great effort to label images correctly~\cite{Richter_2016_ECCV}.
The more convenient alternative is to transfer knowledge from the domain where the labels are generated through computer graphic techniques, like GTA5~\cite{Richter_2016_ECCV} dataset or SYNTHIA ~\cite{7780721} dataset.
However, due to the domain discrepancy between real and synthetic datasets, the hypothesis that the training data and testing data share the same distribution is no longer true and the generalization ability of models trained on synthetic data degrades drastically on data in the target domain.

\begin{figure*}[ht]
\begin{center}
\includegraphics[width=0.95\linewidth]{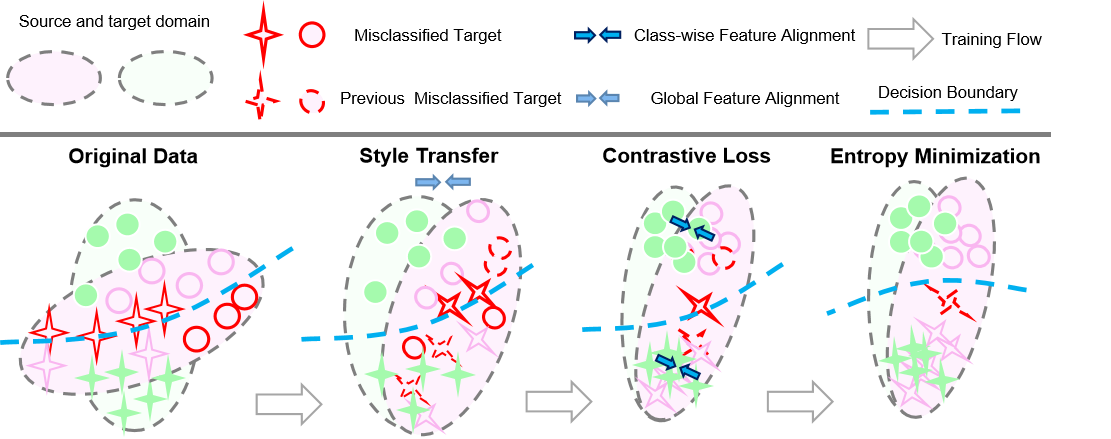}
\end{center}
   \caption{Example of the feature movement in the feature space through CFContra. The original Data part shows massive amounts of target samples are misclassified due to domain discrepancy. Through style transfer, more but not all of the target features can be correctly classified out of global feature alignment. Then by optimizing contrastive loss, features become more concentrated thus easier to distinguish. Through previous misclassified target, we are able to tell the movement of features within each step. Finally, we minimize entropy to move the decision boundary away from features and further increase segmentation accuracy. }
\label{fig:first}
\end{figure*}
  
One of the most important underlying problems of domain discrepancy is the mismatch between feature distributions across domains. There are numerous methods aligning feature distributions in various ways including matching the distribution of input image through style transfer~\cite{DBLP:journals/corr/abs-1712-00479,DBLP:journals/corr/abs-1804-05827, DBLP:journals/corr/abs-1711-03213, DBLP:conf/cvpr/ChangWPC19, DBLP:journals/corr/abs-1812-05418}, the distribution of features via discriminator~\cite{hong2018conditional, ganin2015unsupervised} or losses ~\cite{DBLP:conf/iclr/ShuBNE18}, and distribution of labels through classifiers~\cite{DBLP:conf/iclr/FrenchMF18,DBLP:journals/corr/abs-2001-03182,DBLP:conf/iccv/LianDLG19, DBLP:conf/cvpr/TsaiHSS0C18}.
However, These methods only minimize the global distance between domains without thinking about the mismatch between class-wise features in the target domain and source domain.
There are other algorithms~\cite{DBLP:journals/corr/abs-1910-13049, DBLP:conf/iccv/LeeKKJ19, DBLP:conf/cvpr/Luo0GYY19, DBLP:conf/cvpr/SaitoWUH18, DBLP:conf/cvpr/LeeBBU19} aligning class-wise features using two classifiers based adversarial training or designed losses together with complicated training scheme. The two classifier idea is especially elegant and thought-provoking. 

However, previous class-wise feature alignment algorithms are either too complicated to train or unable to produce robust results due to the adversarial training. 
Therefore, we designed a coarse-to-fine feature alignment method using contrastive loss combined with style transfer.
It is directly trained on transferred images and aligning features through optimizing the contrastive loss without an adversarial network. Therefore, our network is easy to train and experiments show its robustness and state-of-the-art performance. 

We build our method upon the entropy minimization~\cite{Vu_2019_CVPR} algorithm. This algorithm represents the state-of-the-art performance in UDA, and at the same time, provides a powerful performance boost upon the feature alignment method.
The entropy minimization algorithm minimizes the entropy of the predictive vectors to increase the gap between features and the decision boundary. In this way, it improves the model's generalization ability. Combining our method with entropy minimization, we pull the decision boundary further and boost the performance.

Specifically, we first transfer images from the source domain to the target domain using adaptive instance normalization(ADAIN)~\cite{DBLP:journals/corr/HuangB17} based style transfer~\cite{ DBLP:journals/corr/abs-1802-06474}, which is light-weight compared to other style transfer methods~\cite{CycleGAN2017, pix2pix2017}; 
Then we extracted class centers of the source domain and assign pseudo-labels to target features using the transferred images. We assign pseudo-labels based on the hypothesis that features of the same class across domains are closer to each other; Afterwards, we build the memory bank for each category in the source domain and target domain respectively. Therefore we can compare the representative feature center of the whole dataset with features in each image batch and improve the stability of comparison; Finally, we build contrastive loss, entropy loss, and cross-entropy loss for optimization. After optimization, features are concentrate within each class and well separated from features of other categories, thus distinguishable for the classifier.
We present Figure~\ref{fig:first} to illustrate each step's impact on feature alignment and decision boundary.

In particular, contrastive learning in semantic segmentation can be hard to train due to the massive amounts of memory occupation when every pixel counts as an instance. We make the training process viable and improve the model's performance through two tricks: Firstly, we ignore hard pixels in the target domain which is close to several centers in the early stage, and exploit it later when features are drawn closer to the center through the training. Secondly, we update the memory center using the average of class-wise features in the same batch rather than every feature to decrease computation complexity greatly and achieve a huge runtime cutup.

A great number of experiments show that our algorithm outperforms the original entropy minimization algorithm and other state-of-the-art feature alignment methods.~\cite{DBLP:journals/corr/abs-1910-13049, DBLP:journals/corr/abs-1804-05827, DBLP:journals/corr/abs-1711-03213} to a great extend. Also, Experiments show that our coarse-to-fine feature alignment method pulls class-wise features closer and performs better on the target domain than the global feature alignment method or class-wise feature alignment method. Therefore, each of our modules plays an important role in our algorithm. In summary, our main contribution is two-fold and summarized in the following:
\begin{itemize}
    \item Combine contrastive loss and style tranfer in semantic segmentation for the first time. Compared to other class-wise feature alignment methods, our coarse-to-fine feature alignment method is easier to train, more robust, and provides state-of-the-art performance .
    \item Reducing the memory occupation and computational complexity of contrastive learning in semantic segmentation. Through our tricks in building and updating the memory bank, we make contrastive learning feasible and useful in UDA.
\end{itemize}

\section{Related Work}
{\bf Semantic Segmentation}
Semantic segmentation is a pixel-wise classification task used in various applications like autonomous driving, Geo sensing, Precision Agriculture, and so forth. Deep Learning based semantic segmentation has evolved quickly due to the access to large datasets and various network designs ~\cite{DBLP:journals/corr/abs-1802-02611,DBLP:journals/corr/ZhaoSQWJ16,DBLP:journals/corr/abs-1912-08193}.

{\bf Unsupervised Domain Adaptation}
Domain Adaptation(DA) algorithms are a group of algorithms that try to transfer knowledge from one or several sources to a related target, which including resample methods, feature alignment methods, and inference-based methods~\cite{Vu_2019_CVPR}. DA algorithms are of great importance since it helps models to learn new knowledge with ease. In particular, Unsupervised Domain Adaptation transfers knowledge without the need of any new labels, which is especially useful but at the same time challenging.

{\bf Unsupervised Domain Adaptation in Semantic Segmentation}
UDA can be used in various tasks including classification~\cite{DBLP:journals/corr/abs-1909-13589, 10.5555/3157096.3157333}, object detection and so on. However, UDA in semantic segmentation can be extremely difficult. It is easy to align class-wise features in classification tasks since each image belongs to only one class. Nevertheless, we do not know which part of image belongs to a certain class for sure in the semantic segmentation task due to the lack of labels. Therefore, it is impossible to align class-wise features 100\% correct in semantic segmentation though it plays a key role in domain adaptation.

{\bf Feature Alignment Method in UDA}
There are various ways to apply feature alignment methods in UDA, which can be categorized into alignment in image level, feature level, and label level. Various methods~\cite{DBLP:conf/iccv/YueZZSKG19,DBLP:journals/corr/abs-1711-03213,DBLP:journals/corr/abs-1712-00479} use style transfer based on GAN\cite{goodfellow2014generative} and cycle-consistent loss\cite{DBLP:journals/corr/ZhuPIE17} to transfer images from the source domain to the target domain; Other algorithms use maximum mean discrepancy loss or classifier based adversarial training to match distribution at feature level; There are also methods think it is more important to match structural information like the semantic labels. To this end, some algorithms combine several of the above methods and align features in both image-level and feature-level.

{\bf Class-wise Feature Alignment Method in UDA}
It is rather significant yet difficult to align class-wise features in semantic segmentation tasks. Only recently, some methods have been proposed to solve this issue. ~\cite{DBLP:conf/cvpr/SaitoWUH18, DBLP:conf/iccv/LeeKKJ19} tries to use two classifiers with different angles to build reliable boundaries to seperate features in source domain. Combined with adversarial training, the feature extractor will generate target features that lie within the boundary as well, therefore separate features in the target domain. Compared with the above methods, our approach uses contrastive loss without adversarial training or other complex training techniques. Therefore it is easier to train and provides more consistent results.

{\bf Contrastive Learning}
Contrastive learning~\cite{chen2020big,chen2020improved,caron2021unsupervised,chen2020exploring} first used in self supervised learning to train feature extractor without any labels. It controls feature movement based on the principle that positive samples should stay together while negative samples stay apart. Ideally, optimizing the network through contrastive loss will push and pull features in a hypersphere.

{\bf Contrastive Learning in UDA}
Although contrastive learning was proposed only recently, several work~\cite{su2020gradient} have managed to utilize it in the UDA. Due to its effectiveness in matching features, the contrastive loss was used to maximize the mutual information between label and feature~\cite{park2020joint}, minimize intra-class distance, and maximize inter-class distance~\cite{kang2019contrastive}, and so on. However, our approach is the first to our knowledge that uses contrastive learning in semantic segmentation and brings a great boost to the model's performance.

\section{Methods}
In this section, we present the details of our method. Firstly, we formulate the problem and explain the basic setting we used; Secondly, we describe our network architecture the procedures in style transfer and constructing contrastive loss step by step. Finally, we present our objective function.

\subsection{Problem Formulation}
In unsupervised domain adaptation, we have source domain with labels that denoted as $\boldsymbol{D_s} = \{(x_s,y_s) | x_s \subset \mathbb{R}^{H \times W \times 3}, y_s \subset  \mathbb{R}^{H \times W}, y_s \in [1,C]\}$, and we have target domain without labels denoted as $\boldsymbol{D_t} = \{(x_t) | x_t \subset \mathbb{R}^{H \times W \times 3}\}$. With images $x$ input into feature extrator $\boldsymbol{F}$, we get a C-dimensional prediction map after softmax layer: $\boldsymbol{F}(x) = P(x), x \subset  \mathbb{R}^{H \times W \times C}$. For source domain predictions, we constrain it with cross entropy loss written as:

\begin{equation}
\boldsymbol{L}_{CE} = - \sum_{n=1}^{H \times W} \sum_{c=1}^{C} y_s^{n,c} \log P_{x_s}^{n,c}
\label{celoss}
\end{equation}

For target domain predictions, we build up entropy loss descibed as follows.

{\bf Entropy Minimization}
As an effective UDA method, the entropy minimization algorithm minimizes the entropy of $P_x$, which can be viewed as the distribution of predicted results. The entropy loss defines as follows:
\begin{equation}
\boldsymbol{E}_{x_t}^{(h,w)} = -\frac{1}{\log(C)}\sum_{c=1}^{C} P_{x_t}^{h,w,c} \log (P_{x_t}^{h,w,c})
\label{entropyloss}
\end{equation}
Through optimizing the loss, the distribution of predicted result becomes picky and the model is more confident about the output result.
In total, the basic loss is defined as follows with $\lambda_{ent}$ being the weight of entropy loss:
\begin{equation}
\boldsymbol{L}(x_s, x_t) = \frac{1}{|\boldsymbol{D}_s|} \sum_{x_s} \boldsymbol{L}_{CE} + \frac{\lambda_{ent}}{|\boldsymbol{D}_t|} \sum_{x_t} \sum_{h,w} \boldsymbol{E}_{x_t}^{(h,w)}
\end{equation}

\subsection{Coarse-to-fine Feature  Alignment Network}
\begin{figure*}[ht]
\begin{center}
\includegraphics[width=0.9\linewidth]{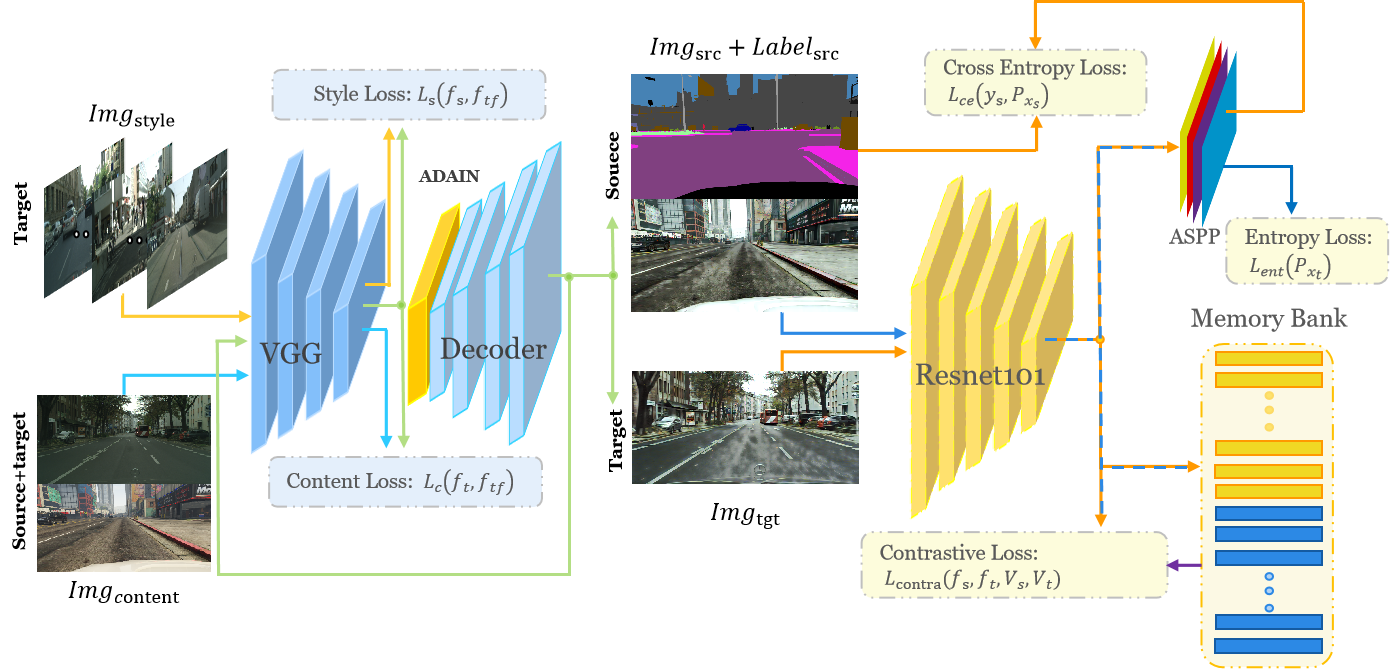}
\end{center}
 \caption{(best viewed in color.) The architecture of our network. Our network combines style transfer with semantic segmentation. First, we transfer images through the pre-trained ADAIN network, and then the transferred images are used to generate components in the contrastive loss. Contrastive loss, cross-entropy loss, and entropy minimization loss constitute our final objective function.}
\label{fig:long}
\end{figure*}

The overall network architecture of our approach is based on autoencoder~\cite{Ballard1987ModularLI} and Deeplabv2~\cite{CP2016Deeplab} and is shown in Figure~\ref{fig:long}. 

\subsubsection{Coarse Feature Alignment via Style Transfer}
We transfer both source domain images and target domain images to the source domain using pre-trained ADAIN based style transfer network~\cite{DBLP:journals/corr/HuangB17}, which view mean and variance of images as style.  In the style transfer network, we extract source domain features $f_s$, target domain features $f_t$, and map the mean and variance of sourece domain features to target domain features. Through style transfer, we align global features across domains. The process in the ADAIN module can be written as:

\begin{equation}
    f_s = \frac{f_s-\mu_s}{\sqrt{\sigma_s}}
\end{equation}
\begin{equation}
    f_s = f_s*\sqrt{\sigma_t} + \mu_t
\end{equation}
The $\mu_s$,$\mu_t$ and $\sigma_s$,$\sigma_t$ is the mean and the variance of the source domain and the target domain respectively.

To restrain the style and content of the generated image, we  train the network using the following content loss and style loss with $Img_{tf}$ as the generated image and $f_{tf}$ as its feature:

\begin{equation}
\boldsymbol{L}_{content} = \frac{1}{H\times W \times C } \sum_{n=1}^{H \times W}\sum_{c=1}^{C}(f_{tf}^{n, c}-f_s^{n, c})^2
\end{equation}

\begin{equation}
    L_{style} = \frac{1}{2} ((\mu_{tf}-\mu_t)^2 + (\sqrt{\sigma_t}-\sqrt{\sigma_{tf}})^2)
\end{equation}

The style transfer network is trained without adversarial losses and based on VGG and a decoder, which is easier to train and light-weight compared to other state-of-the-art style transfer network.~\cite{CycleGAN2017, pix2pix2017}

\subsubsection{Class-wise Feature Alignment via Contrastive Loss}
Aligning class-wise features plays a vital role in domain adaptation since it's more accurate and helps the model perform better than global feature alignment. Previous work\cite{DBLP:journals/corr/abs-1911-05722} proves that contrastive loss is useful in clustering features. Therefore, we apply the contrastive loss to cluster the class-wise target features and their corresponding source features. We use the InfoNCE~\cite{DBLP:journals/corr/abs-1807-03748} with the similarity measure function being the inner product function as our contrastive loss:
\begin{equation}
    \boldsymbol{L}_{contra}^i=-\log \frac{\exp( <(f(x_i),f(x^+))>/ \tau)}{\sum_{k\ne i}^{N} \exp (<f(x_{i}), f(x_{k})> / \tau)}
\label{contraloss}
\end{equation}

$f(x_i)$ represents the feature and $f(x^+)$ is the positive center it needs to align, whereas $f(x_{k})$ represents the negative centers it needs to be drawn away from. And $\tau$ is the temperature parameter. Through optimizing the contrastive loss, features will be drawn to the center of its label or its pseudo label indicates. 
Key procedures in constructing contrastive loss include remapping features, pseudo-label assignment, and constructing the memory bank.

{\bf Remapping Features in Contrastive Loss}
Inspired by other work in contrastive learning~\cite{grill2020bootstrap, chen2020simple, DBLP:journals/corr/abs-1911-05722}, it is important to remap the features used in semantic segmentation and decouple features' functionality. Therefore, we plug in head modules that contain different layers after the resnet~\cite{DBLP:journals/corr/HeZRS15} backbone. Thus, segmentation features are different with features used in the contrastive loss. Different design is borrowed from ~\cite{grill2020bootstrap, chen2020simple, DBLP:journals/corr/abs-1911-05722} and described in Table~\ref{ffc}.

\begin{table}[ht]
\begin{center}
\resizebox{0.47\textwidth}{!}{
\begin{tabular}{ c|c|c|c|c  }
\hline
Module Name & Linear & MOCO~\cite{DBLP:journals/corr/abs-1911-05722} & BYOL~\cite{grill2020bootstrap} & SIMCLR~\cite{chen2020simple}\\
\hline
Structure & Linear & Linear  & Linear  & Linear \\
& & RELU & BatchNorm1d & BatchNorm1d\\
& & Linear & RELU & RELU\\
& & & Linear & Linear \\
& & & & BatchNorm1d \\
\hline
\end{tabular}}
\end{center}
\caption{Different structure of head module}
\label{ffc}
\end{table}

{\bf Pseudo-label Assignment}
Since we do not have labels in the target domain, each target domain feature will be assigned a pseudo label based on its distance to other source centers.
With the hypothesis that features in the same category are close to each other, the label of each target domain feature is the index of source center which the feature is closest to. Furthermore, to increase label accuracy, we ignore features that do not have enough distance differences between source centers. Each center $V_s^i$ and $V_t^i$ are represented by the mean of features in each category of source and target domain, such that we can stabilize the comparing process. And each source center is calculated as:
\begin{equation}
    V_s^i = \frac{1}{\sum_{n=1}^{H \times W}I_{y_s^n=i}(y_s^n)}\sum_{n=1}^{H \times W}I_{y_s^n=i}(y_s^n)*f_s^n
\end{equation}
where $I_{y_s^n=i}(y_s^n)$ is the indication function and equals 1 only when $y_s^n=i$ otherwise equals 0. It will be abbreviated as $I_{y_s^n=i}$ in the following out of convenience. And $f_s^n$ represents the source domain features extracted by the resnet~\cite{DBLP:journals/corr/HeZRS15} backbone. Therefore, source centers are calculated by averaging all features in the source domain by category.

Then we calculate the distance between the target features and the source centers.  First, we calculate the minimum distance with $f_t^n$ represents the target domain features:
\begin{equation}
    d_{min}^{n} = \min_{i}(\sqrt{(f_t^{n}-V_s^{i})})
\end{equation}
Then we calculate the second minimum distance with the feature's closest center indexed by $k$:
\begin{equation}
    d_{secmin}^{n} = \min_{i\ne k}(\sqrt{(f_t^{n}-V_s^{i})})
\end{equation}

At last, the target feature will be assigned label $k$ if the difference between distances is larger than a threshold $t$. And we can calculate target centers based on the pseudo labels.
\begin{equation}
y_t^{n} =
\begin{cases}
k& |d_{min}^{n}-d_{secmin}^n| > t\\
-1& \text{otherwise}
\end{cases}
\end{equation}

\begin{equation}
V_t^i = \frac{1}{\sum_{n=1}^{H \times W}I_{y_t^n=i}}\sum_{n=1}^{H \times W}I_{y_t^n=i}*f_t^n
\end{equation}

Note that features with label -1 will not be saved in the memory bank and will be ignored by contrastive loss. In this way, we can save lots of memory occupation of memory bank and increased pseudo-label accuracy. As training proceeds, features will be pulled closer to its center and assigned labels, such that it can be utilized in contrastive loss.

{\bf Construct Memory Bank}
With the centers calculated above, our memory bank is built and we use momentum $\alpha$ to update the memory bank. With the memory bank, we can compare features of each batch with global average features. And updating the memory bank with momentum moves centers slowly and stabilizes the clustering process.
Furthermore, we update the center using the average of each batch's data rather than all of the features in each batch, therefore we can reduce computational complexity. The updating process can be written as follows.

First, we calculate the mean $M$ of features in source domain and target domaineach by category with $B$ being the batch size:
\begin{equation}
M_s^i = \frac{1}{\sum_{n=1}^{H\times W \times B} I_{y_s^n=i}}\sum_{n=1}^{H\times W \times B} I_{y_s^n=i} \times f_s^{n}
\end{equation}

\begin{equation}
M_t^i = \frac{1}{\sum_{n=1}^{H\times W \times B} I_{y_t^n=i}}\sum_{n=1}^{H\times W \times B} I_{y_t^n=i} \times f_t^{n}
\end{equation}

Then, we update the memory bank as follows with $\alpha$ represents the momentum:
\begin{equation}
    V_s^i = \alpha V_s^i + (1-\alpha) M_s^i
\end{equation}
\begin{equation}
    V_t^i = \alpha V_t^i + (1-\alpha) M_t^i
\end{equation}

Normally, we will update the center with every feature in the batch, rather than the average of it. Through the above approximation, we make the training process much more efficient without harming the performance.

{\bf Contrative loss}
With the obtained target center and the source center, our contrastive loss within source domain can be written as:

\begin{equation}
\boldsymbol{L}_{contra}^i(f_s, V_s)=-\log \frac{\frac{\exp(<f_s^n,V_s^+>)}{\tau}}{\sum_{i\ne k}^{N} \frac{\exp(<f_s^n,V_s^i>)}{\tau}}
\label{contralossnew}
\end{equation}

where $V_s^+$ is the corresponding center $k$ of current feature.

Within the contrastive loss, comparison can intertwine between the source domain and target domain, the contrastive loss that achieved the best performance is:
\begin{equation}
\begin{aligned}
\boldsymbol{L}_{contra}^i\!&=\! \boldsymbol{L}_{contra}^i(f_s, V_s) +\boldsymbol{L}_{contra}^i(f_s, V_t) \\&+ \boldsymbol{L}_{contra}^i(f_t, V_s)+\boldsymbol{L}_{contra}^i(f_t, V_t)
\end{aligned}
\label{contralossend}
\end{equation}

The overall objective function for each iteration with $B_s$ and $B_t$ being the batch size of the source domain and target domain is written as:
\begin{equation}
\begin{aligned}
\boldsymbol{L}(x_s, x_t) &= \frac{1}{|B_s|} \sum_{B_s} \boldsymbol{L}_{CE} \\ &+ \frac{\lambda_{ent}}{|B_t|} \sum_{B_t} \sum_{h,w} \boldsymbol{E}_{x_t}^{(h,w)} \\&+ \frac{\lambda_{contra}}{|B_t|} \sum_{B_t} \sum_{h,w} \boldsymbol{L}_{contra}^{(h,w)}
\end{aligned}
\end{equation}

In total, we train our coarse-to-fine feature alignment network with cross-entropy loss defined in Eq.~\ref{celoss} to build an accurate decision boundary on the transferred source domain. Then we align features with style transfer and contrastive loss defined in Eq.~\ref{contralossend}. Finally, entropy loss defined in Eq.~\ref{entropyloss} helps to broaden the gap between the features and the interface.

\section{Experiments}
In this section, we first compared our algorithm with the MinEnt algorithm~\cite{Vu_2019_CVPR} we build on, as well as other state-of-the-art feature alignment method. Comparison shows the effectiveness of our algorithm. Then, we use an ablation study to find out the role of each module in our algorithm. At last, we dig into the contrastive loss. We firstly show its clustering effect through the change of losses and pseudo-label accuracy, then we analyze its sensitivity to all kinds of parameters, and finally, analyze the effect of different head modules.

\begin{table*}[ht]
\resizebox{1.0\textwidth}{!}{
\begin{tabular}{ c|cccccccccccccccccccc  }
\hline
Method & \rotatebox{90}{road} & \rotatebox{90}{sdwk} & \rotatebox{90}{bldg} & \rotatebox{90}{wall} & \rotatebox{90}{fence} & \rotatebox{90}{pole} & \rotatebox{90}{light} & \rotatebox{90}{sign} & \rotatebox{90}{vege.} & \rotatebox{90}{ter.} & \rotatebox{90}{sky} & \rotatebox{90}{pers.} & \rotatebox{90}{rider} & \rotatebox{90}{car} & \rotatebox{90}{truck} & \rotatebox{90}{bus} & \rotatebox{90}{train} & \rotatebox{90}{moto.} & \rotatebox{90}{bike} & mIOU  \\  \hline
baseline & 60.1 & 20.7 & 66.9 & 14.2 & 21.1 & 26 & 30.2 & 20.7 & 78.4 & 8 & 72.6 & 53.9 & 27.3 & 73.6 & 26.4 & 4.9 & 0 & 25.5 & 34.2 & 35 \\
AdaptSegNet~\cite{DBLP:conf/cvpr/TsaiHSS0C18} & 86.5 & 25.9 & 79.8 & 22.1 & 20 & 23.6 & 33.1 & 21.8 & 81.8 & 25.9 & 75.8 & 57.3 & 26.2 & 76.3 & 29.3 & 32.1 & 7.2 & 29.5 & 32.5 & 41.4 \\
DCAN~\cite{DBLP:journals/corr/abs-1804-05827} & 85 & 30.8 & 81.3 & 25.8 & 21.2 & 22.2 & 25.4 & 26.6 & 83.4 & \textbf{36.7} & 76.2 & 58.9 & 24.9 & 80.7 & 29.5 & \textbf{42.9} & 2.5 & 26.9 & 11.6 & 41.7 \\
Cycada~\cite{DBLP:journals/corr/abs-1711-03213} & 86.7 & 35.6 & 80.1 & 19.8 & 17.5 & \textbf{38} & \textbf{39.9} & \textbf{41.5} & 82.7 & 27.9 & 73.6 & \textbf{64.9} & 19 & 65 & 12 & 28.6 & 4.5 & 31.1 & 42 & 42.7 \\
CLAN~\cite{DBLP:journals/corr/abs-1910-13049} & 87 & 27.1 & 79.6 & \textbf{27.3} & 23.3 & 28.3 & 35.5 & 24.2 & 83.6 & 27.4 & 74.2 & 58.6 & 28 & 76.2 & 33.1 & 36.7 & \textbf{6.7} & \textbf{31.9} & 31.4 & 43.2 \\
MinEnt~\cite{Vu_2019_CVPR} & 86.2 & 18.6 & 80.3 & 27.2 & 24 & 23.4 & 33.5 & 24.7 & 83.8 & 31 & 75.6 & 54.6 & 25.6 & 85.2 & 30 & 10.9 & 0.1 & 21.9 & 37.1 & 42.3 \\
\textbf{CFContra (ours)}  & \textbf{89.7} & \textbf{41.1} & \textbf{82.6} & 23.1 & \textbf{25.2} & 27.5 & 37.2 & 21.8 & \textbf{83} & 35.7 & \textbf{80.9} & 60.3 & \textbf{28.8} & \textbf{85.8} & \textbf{33.9} & 39.2 & 2.3 & 28.7 & \textbf{43.7} & \textbf{45.8} \\ \hline
\end{tabular}}
\caption{Results on GTA5-to-Cityscapes experiment. All of the algorithm above based on deeplabv2~\cite{CP2016Deeplab} with resnet101~\cite{DBLP:journals/corr/HeZRS15}  backbone and CFContra is our coarse-to-fine feature alignment using contrastive loss }
\label{overallres}
\end{table*}

\subsection{Datasets}
We evaluate our model on the  GTA5~\cite{Richter_2016_ECCV} dataset to the Cityscapes~\cite{Cordts2016Cityscapes} dataset.
The GTA5 dataset contains 24966 images with 33 categories. We use 19 classes which is in common with the categories in Cityscapes and all of the images as the training dataset.
The cityscapes dataset contains 2975 images, we use its original training and validation set as training and test set.

\subsection{Implementation Details}
During style transfer, we scale the images
into (512,1024), and train the network for 160000 iterations with the learning rate equals to 1e-4. We use the model of iteration 160000 as the inference model to generate transferred images.

As for contrastive learning, we scale both the source domain and target domain images into (1280, 640) and train the network with the learning rate equals to 2.5e-4 for 120000 iterations. We set the batch size equals to 1 due to memory limitations, and we set the efficient of cross-entropy loss, entropy loss and contrastive loss as 1, 1e-3, 1e-3 respectively. For contrastive loss, the best result is produced when the threshold, the temperature coefficient, and the momentum for the memory bank are 0.05,0.07,0.9 respectively. 
Experiments are conducted using the model pre-trained on Imagenet~\cite{5206848} and the algorithm is implemented using Pytorch~\cite{DBLP:journals/corr/abs-1912-01703} on a single TITAN rtx.

\subsection{Main Results}

\subsubsection{Overall Results}
We compared our algorithm with other state-of-the-art algorithms with the same network structure in Table~\ref{overallres}.
As shown in the table, our method improved the MinEnt method by a large margin, and exceeds other state-of-the-art method like AdaptSegNet~\cite{DBLP:conf/cvpr/TsaiHSS0C18}, DCAN~\cite{DBLP:journals/corr/abs-1804-05827}, Cycada~\cite{DBLP:journals/corr/abs-1711-03213} and CLAN~\cite{DBLP:journals/corr/abs-1910-13049} to a great extent as well. Especially, CLAN is a state-of-the-art method in aligning class-wise features, which shows our algorithm's effectiveness. Compared to other algorithms in class-wise IOU, our algorithms perform better on most classes like road, sidewalk, buildings, fence, vegetation, sky, rider, car, truck, and bike. It illustrates that our CFContra methods help most e classes cluster features no matter how many pixels each class has, and therefore improved accuracy.

\begin{figure*}[ht]
\begin{center}
\includegraphics[width=0.98\linewidth]{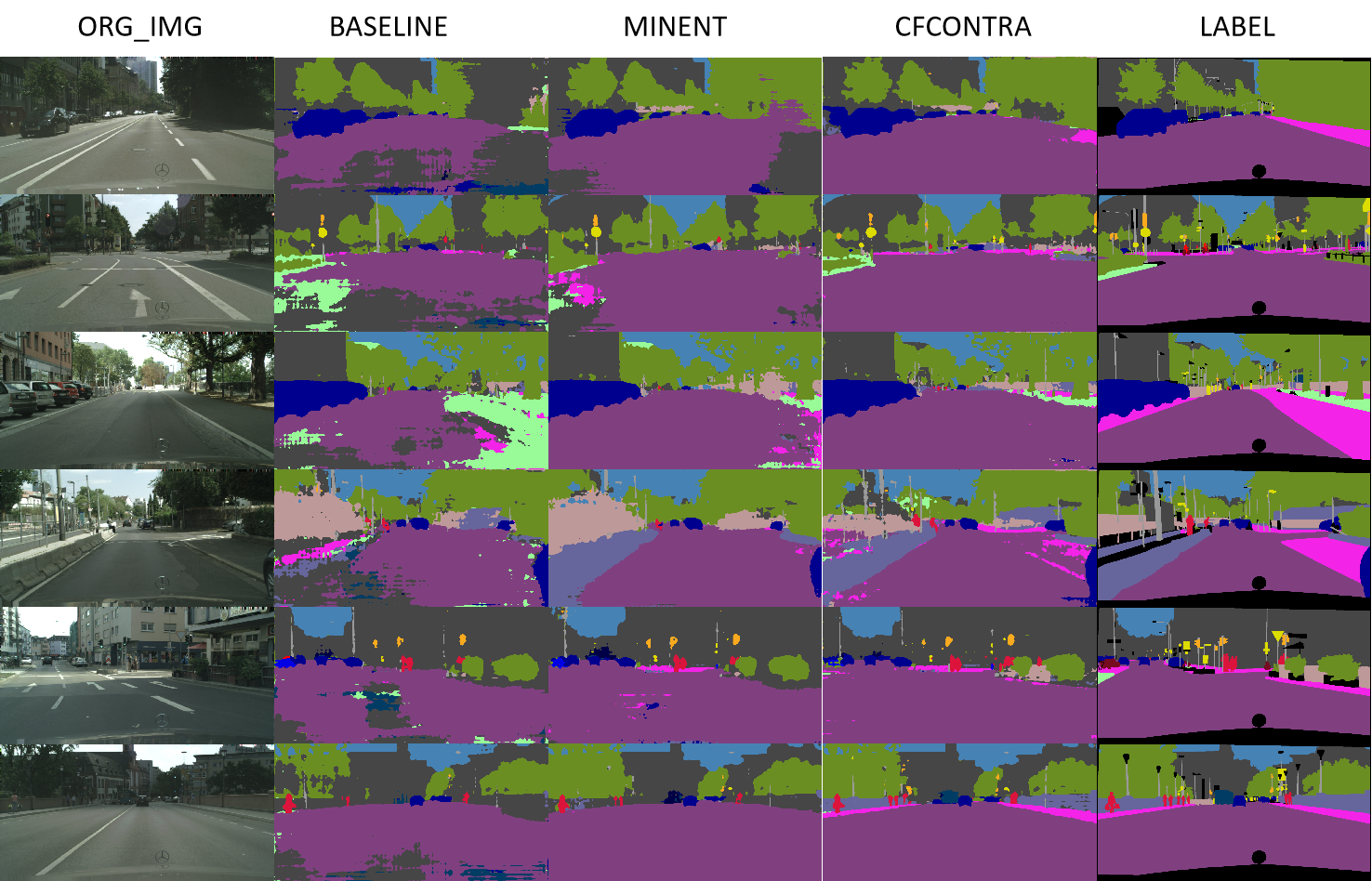}
\end{center}
   \caption{The segmentation results in the target domain. We compared our algorithm with MinEnt\cite{Vu_2019_CVPR} method and direct adaptation. The result shows that through feature alignment, the object edge is refined and prediction within an instance is more consistent.}
\label{picres}
\end{figure*}

\subsubsection{Ablation Study}
As shown in the table~\ref{ablationres}, Adding the style transfer module or adding the contrastive loss alone on top of the MinEnt Algorithm can boost the performance to a great extend. It shows the benefit of both global feature alignment methods. However, simply combine those two feature alignment methods can further boost adaptation performance. This result validates our algorithm and shows the importance of aligning class-wise features on top of aligning global features.

We also present the segementation result on the same picture with different methods in \ref{picres}. The comparison shows that through feature alignment, predictions are more clear and more objects can be identified.

\begin{table}[ht]
\resizebox{0.47\textwidth}{!}{
\begin{tabular}{ ccc|c}
\hline
Entropy & Style transfer & Contrastive loss & mIOU \\  \hline
$\surd$ & & & 42.3 \\
$\surd$ & $\surd$ & & 44.8 \\
$\surd$ & &$\surd$ & 45.4 \\
$\surd$ &$\surd$ &$\surd$ & \textbf{45.8}\\ \hline
\end{tabular}}
\caption{Ablation study}
\label{ablationres}
\end{table}

\subsubsection{Analysis on Contrastive Loss}
Previously, we analyze contrastive loss's clustering ability through Eq.~\ref{contraloss}. With experiment, more evidences rise up and validate our interpretation.
In the Table~\ref{lossandacc}, we listed the change of cross entropy loss, entropy loss and pseudo label accuracy with and without contrastive loss.

As shown in the table, the contrastive loss can further reduce both cross-entropy loss and entropy loss. Cross-entropy loss reflects the accuracy of source domain prediction and entropy loss shows the distance between decision boundary and feature clusters. Furthermore, the pseudo label accuracy, which indicates the percentage of features that have been assigned the correct center, also improves. These shreds of evidence show that by optimizing the contrastive loss we construct, we can bring both source and target domain features close to their corresponding center.

\begin{table}[ht]
\resizebox{0.47\textwidth}{!}{
\begin{tabular}{ c|ccc}
\hline
Method & CE loss & Entropy loss & Pseu-label acc \\
\hline
baseline & 0.160 & - & - \\
MinEnt & 0.148 & 0.148 & - \\ \hline
CONTRA(intra) & 0.128 & 0.104 & +5\% \\
CONTRA(all)  & 0.115 & \textbf{0.08} &  +3\%  \\
CONTRA(MOCO) & \textbf{0.112 }& 0.09 &  \textbf{+15\%} \\ \hline
\end{tabular}}
\caption{Change of losses and pseudo-label accuracy with contrastive loss. CONTRA represents we only align class-wise features in the experiment. CONTRA(intra) represents we only cluster features within each domain; CONTRA(all) represents we cluster features within and across domain; CONTRA(MOCO) represents we use MOCO as the head module and cluster features across domains.}
\label{lossandacc}
\end{table}

{\bf Parameter Sensitivity Analysis}
We shows the sensitivity of contrastive loss to parameters like $\tau$, $\alpha$, $t$, and $\lambda_{contra}$ in the Table~\ref{sensitivityres}. 

Both the temperature coefficient $\tau$ and weight coefficient $\lambda_{contra}$ have a great impact on the performance, whereas the momentum $\alpha$ and threshold $t$ only change the results slightly.
With the contrastive coefficient being too large, the network may overfocus on contrasive loss and move features incorrectly since the pseudo-label is not accurate in the early stage.
With in the range we specified in the Tabel~\ref{sensitivityres}, the temperature coefficient is the bigger the better. According to~\cite{chen2020intriguing}, increasing the temperature coefficient concentate the network on the dissimilarity between negative samples and the distribution differences between negative samples and uniform distribution. By increasing it within the range, We draw negative samples further away from each other thus easier for classifier to distinguish.
The results of momentum and threshold ablation experiments shows that the contrastive loss is robust via the shift in these two parameters within the given range.

\begin{table}[ht]
\begin{center}
\resizebox{0.47\textwidth}{!}{
\begin{tabular}{ ccccc}
\hline
parameter $\tau$ = & 0.03 & 0.05 & 0.07 & 0.14 \\
mIOU & 43.2 & 44.3 & \textbf{44.4} & 44.6 \\ \hline
parameter $\alpha$ =  & 0.9 & 0.99 & &   \\
mIOU & \textbf{44.5} & 44.4 &  &  \\ \hline
parameter threshold $t$ = & 0.03 & 0.05 & &  \\
mIOU & 44.5 & \textbf{44.5} &  &  \\ \hline
parameter $\lambda_{contra}$ =  & 0.1 & 0.01 & 0.001&   \\
mIOU & 31.6 & 41.3 & \textbf{44.5} &  \\ \hline
\end{tabular}}
\end{center}
\caption{Parameter sensitivity. Note that the experiment conducted with only class-wise feature alignment and the result slight droped compared to the whole algorithm.}
\label{sensitivityres}
\end{table}

{\bf Effect of Remapping}
We compared different head modules' effectiveness in our method in Table~\ref{headres}. All of the variants of the head module besides linear module is useful in boosting the performance. The head module named MOCO boost the performance to the greatest extent. It shows that decoupling the features for different tasks is necessary and useful. Since the linear module only scales the features in a given space, it is useless in decoupling therefore useless in improve performance.

\begin{table}[ht]
\resizebox{0.47\textwidth}{!}{
\begin{tabular}{ c|ccccc }
\hline
Head Module & None & Linear & SIMCLR & BYOL & \textbf{MOCO} \\ 
mIOU  & 44.4  & 44.0  & 44.8  & 44.9  & \textbf{45.4} \\
\hline
\end{tabular}}
\caption{Results on different head layer with class-wise alignment only. None represent we directly use the feature after backbone for both semantic segmentation and contrastive loss. Other variant use the head module its name indicates.}
\label{headres}
\end{table}

\section{Conclusion}
In this paper, we propose coarse-to-fine feature alignment using contrastive loss called as CFContra in the UDA for semantic segmentation. Compared with other work aligning class-wise features, our method does not need adversarial training or other complicated training scheme, therefore is easy to train and experiments produce robust, state-of-the-art results. Also, we improved both time and space efficiency of contrastive learning through tricks in constructing and updating the memory bank. Furthermore, we analyze the contrastive loss in various aspects and further understands it clustering ability.

{\small
\bibliographystyle{ieee_fullname}
\bibliography{egbib}

\begin{thebibliography}{10}\itemsep=-1pt

\bibitem{Ballard1987ModularLI}
D. Ballard.
\newblock Modular learning in neural networks.
\newblock In {\em AAAI}, 1987.

\bibitem{caron2021unsupervised}
Mathilde Caron, Ishan Misra, Julien Mairal, Priya Goyal, Piotr Bojanowski, and
  Armand Joulin.
\newblock Unsupervised learning of visual features by contrasting cluster
  assignments, 2021.

\bibitem{DBLP:conf/cvpr/ChangWPC19}
Wei{-}Lun Chang, Hui{-}Po Wang, Wen{-}Hsiao Peng, and Wei{-}Chen Chiu.
\newblock All about structure: Adapting structural information across domains
  for boosting semantic segmentation.
\newblock In {\em {IEEE} Conference on Computer Vision and Pattern Recognition,
  {CVPR} 2019, Long Beach, CA, USA, June 16-20, 2019}, pages 1900--1909.
  Computer Vision Foundation / {IEEE}, 2019.

\bibitem{DBLP:journals/corr/ChenPSA17}
Liang{-}Chieh Chen, George Papandreou, Florian Schroff, and Hartwig Adam.
\newblock Rethinking atrous convolution for semantic image segmentation.
\newblock {\em CoRR}, abs/1706.05587, 2017.

\bibitem{DBLP:journals/corr/abs-1802-02611}
Liang{-}Chieh Chen, Yukun Zhu, George Papandreou, Florian Schroff, and Hartwig
  Adam.
\newblock Encoder-decoder with atrous separable convolution for semantic image
  segmentation.
\newblock {\em CoRR}, abs/1802.02611, 2018.

\bibitem{CP2016Deeplab}
Liang-Chieh Chen, George Papandreou, Iasonas Kokkinos, Kevin Murphy, and Alan~L
  Yuille.
\newblock Deeplab: Semantic image segmentation with deep convolutional nets,
  atrous convolution, and fully connected crfs.
\newblock {\em arXiv:1606.00915}, 2016.

\bibitem{DBLP:journals/corr/abs-1909-13589}
Minghao Chen, Hongyang Xue, and Deng Cai.
\newblock Domain adaptation for semantic segmentation with maximum squares
  loss.
\newblock {\em CoRR}, abs/1909.13589, 2019.

\bibitem{chen2020simple}
Ting Chen, Simon Kornblith, Mohammad Norouzi, and Geoffrey Hinton.
\newblock A simple framework for contrastive learning of visual
  representations.
\newblock 2020.

\bibitem{chen2020big}
Ting Chen, Simon Kornblith, Kevin Swersky, Mohammad Norouzi, and Geoffrey
  Hinton.
\newblock Big self-supervised models are strong semi-supervised learners, 2020.

\bibitem{chen2020intriguing}
Ting Chen and Lala Li.
\newblock Intriguing properties of contrastive losses.
\newblock 2020.

\bibitem{chen2020improved}
Xinlei Chen, Haoqi Fan, Ross Girshick, and Kaiming He.
\newblock Improved baselines with momentum contrastive learning.
\newblock 2020.

\bibitem{chen2020exploring}
Xinlei Chen and Kaiming He.
\newblock Exploring simple siamese representation learning, 2020.

\bibitem{DBLP:journals/corr/abs-2001-03182}
Yun{-}Chun Chen, Yen{-}Yu Lin, Ming{-}Hsuan Yang, and Jia{-}Bin Huang.
\newblock Crdoco: Pixel-level domain transfer with cross-domain consistency.
\newblock {\em CoRR}, abs/2001.03182, 2020.

\bibitem{Cordts2016Cityscapes}
Marius Cordts, Mohamed Omran, Sebastian Ramos, Timo Rehfeld, Markus Enzweiler,
  Rodrigo Benenson, Uwe Franke, Stefan Roth, and Bernt Schiele.
\newblock The cityscapes dataset for semantic urban scene understanding.
\newblock In {\em Proc. of the IEEE Conference on Computer Vision and Pattern
  Recognition (CVPR)}, 2016.

\bibitem{5206848}
J. {Deng}, W. {Dong}, R. {Socher}, L. {Li}, {Kai Li}, and {Li Fei-Fei}.
\newblock Imagenet: A large-scale hierarchical image database.
\newblock In {\em 2009 IEEE Conference on Computer Vision and Pattern
  Recognition}, pages 248--255, 2009.

\bibitem{pascal-voc-2012}
M. Everingham, L. Van~Gool, C.~K.~I. Williams, J. Winn, and A. Zisserman.
\newblock The {PASCAL} {V}isual {O}bject {C}lasses {C}hallenge 2012 {(VOC2012)}
  {R}esults.
\newblock
  http://www.pascal-network.org/challenges/VOC/voc2012/workshop/index.html.

\bibitem{DBLP:conf/iclr/FrenchMF18}
Geoffrey French, Michal Mackiewicz, and Mark~H. Fisher.
\newblock Self-ensembling for visual domain adaptation.
\newblock In {\em 6th International Conference on Learning Representations,
  {ICLR} 2018, Vancouver, BC, Canada, April 30 - May 3, 2018, Conference Track
  Proceedings}. OpenReview.net, 2018.

\bibitem{ganin2015unsupervised}
Yaroslav Ganin and Victor Lempitsky.
\newblock Unsupervised domain adaptation by backpropagation.
\newblock 2015.

\bibitem{DBLP:journals/corr/abs-1812-05418}
Rui Gong, Wen Li, Yuhua Chen, and Luc~Van Gool.
\newblock {DLOW:} domain flow for adaptation and generalization.
\newblock {\em CoRR}, abs/1812.05418, 2018.

\bibitem{goodfellow2014generative}
Ian~J. Goodfellow, Jean Pouget-Abadie, Mehdi Mirza, Bing Xu, David
  Warde-Farley, Sherjil Ozair, Aaron Courville, and Yoshua Bengio.
\newblock Generative adversarial networks.
\newblock 2014.

\bibitem{grill2020bootstrap}
Jean-Bastien Grill, Florian Strub, Florent Altché, Corentin Tallec, Pierre~H.
  Richemond, Elena Buchatskaya, Carl Doersch, Bernardo~Avila Pires,
  Zhaohan~Daniel Guo, Mohammad~Gheshlaghi Azar, Bilal Piot, Koray Kavukcuoglu,
  Rémi Munos, and Michal Valko.
\newblock Bootstrap your own latent: A new approach to self-supervised
  learning.
\newblock 2020.

\bibitem{DBLP:journals/corr/abs-1911-05722}
Kaiming He, Haoqi Fan, Yuxin Wu, Saining Xie, and Ross~B. Girshick.
\newblock Momentum contrast for unsupervised visual representation learning.
\newblock {\em CoRR}, abs/1911.05722, 2019.

\bibitem{DBLP:journals/corr/HeZRS15}
Kaiming He, Xiangyu Zhang, Shaoqing Ren, and Jian Sun.
\newblock Deep residual learning for image recognition.
\newblock {\em CoRR}, abs/1512.03385, 2015.

\bibitem{DBLP:journals/corr/abs-1711-03213}
Judy Hoffman, Eric Tzeng, Taesung Park, Jun{-}Yan Zhu, Phillip Isola, Kate
  Saenko, Alexei~A. Efros, and Trevor Darrell.
\newblock Cycada: Cycle-consistent adversarial domain adaptation.
\newblock {\em CoRR}, abs/1711.03213, 2017.

\bibitem{hong2018conditional}
Weixiang Hong, Zhenzhen Wang, Ming Yang, and Junsong Yuan.
\newblock Conditional generative adversarial network for structured domain
  adaptation.
\newblock In {\em Proceedings of the IEEE Conference on Computer Vision and
  Pattern Recognition}, pages 1335--1344, 2018.

\bibitem{DBLP:journals/corr/HuangB17}
Xun Huang and Serge~J. Belongie.
\newblock Arbitrary style transfer in real-time with adaptive instance
  normalization.
\newblock {\em CoRR}, abs/1703.06868, 2017.

\bibitem{pix2pix2017}
Phillip Isola, Jun-Yan Zhu, Tinghui Zhou, and Alexei~A Efros.
\newblock Image-to-image translation with conditional adversarial networks.
\newblock {\em CVPR}, 2017.

\bibitem{kang2019contrastive}
Guoliang Kang, Lu Jiang, Yi Yang, and Alexander~G Hauptmann.
\newblock Contrastive adaptation network for unsupervised domain adaptation,
  2019.

\bibitem{DBLP:journals/corr/abs-1912-08193}
Alexander Kirillov, Yuxin Wu, Kaiming He, and Ross~B. Girshick.
\newblock Pointrend: Image segmentation as rendering.
\newblock {\em CoRR}, abs/1912.08193, 2019.

\bibitem{DBLP:conf/cvpr/LeeBBU19}
Chen{-}Yu Lee, Tanmay Batra, Mohammad~Haris Baig, and Daniel Ulbricht.
\newblock Sliced wasserstein discrepancy for unsupervised domain adaptation.
\newblock In {\em {IEEE} Conference on Computer Vision and Pattern Recognition,
  {CVPR} 2019, Long Beach, CA, USA, June 16-20, 2019}, pages 10285--10295.
  Computer Vision Foundation / {IEEE}, 2019.

\bibitem{DBLP:conf/iccv/LeeKKJ19}
Seungmin Lee, Dongwan Kim, Namil Kim, and Seong{-}Gyun Jeong.
\newblock Drop to adapt: Learning discriminative features for unsupervised
  domain adaptation.
\newblock In {\em 2019 {IEEE/CVF} International Conference on Computer Vision,
  {ICCV} 2019, Seoul, Korea (South), October 27 - November 2, 2019}, pages
  91--100. {IEEE}, 2019.

\bibitem{DBLP:journals/corr/abs-1802-06474}
Yijun Li, Ming{-}Yu Liu, Xueting Li, Ming{-}Hsuan Yang, and Jan Kautz.
\newblock A closed-form solution to photorealistic image stylization.
\newblock {\em CoRR}, abs/1802.06474, 2018.

\bibitem{DBLP:conf/iccv/LianDLG19}
Qing Lian, Lixin Duan, Fengmao Lv, and Boqing Gong.
\newblock Constructing self-motivated pyramid curriculums for cross-domain
  semantic segmentation: {A} non-adversarial approach.
\newblock In {\em 2019 {IEEE/CVF} International Conference on Computer Vision,
  {ICCV} 2019, Seoul, Korea (South), October 27 - November 2, 2019}, pages
  6757--6766. {IEEE}, 2019.

\bibitem{lin2014microsoft}
Tsung-Yi Lin, Michael Maire, Serge Belongie, Lubomir Bourdev, Ross Girshick,
  James Hays, Pietro Perona, Deva Ramanan, C.~Lawrence Zitnick, and Piotr
  Dollár.
\newblock Microsoft coco: Common objects in context, 2014.
\newblock cite arxiv:1405.0312Comment: 1) updated annotation pipeline
  description and figures; 2) added new section describing datasets splits; 3)
  updated author list.

\bibitem{DBLP:conf/cvpr/Luo0GYY19}
Yawei Luo, Liang Zheng, Tao Guan, Junqing Yu, and Yi Yang.
\newblock Taking a closer look at domain shift: Category-level adversaries for
  semantics consistent domain adaptation.
\newblock In {\em {IEEE} Conference on Computer Vision and Pattern Recognition,
  {CVPR} 2019, Long Beach, CA, USA, June 16-20, 2019}, pages 2507--2516.
  Computer Vision Foundation / {IEEE}, 2019.

\bibitem{DBLP:journals/corr/abs-1712-00479}
Zak Murez, Soheil Kolouri, David~J. Kriegman, Ravi Ramamoorthi, and Kyungnam
  Kim.
\newblock Image to image translation for domain adaptation.
\newblock {\em CoRR}, abs/1712.00479, 2017.

\bibitem{park2020joint}
Changhwa Park, Jonghyun Lee, Jaeyoon Yoo, Minhoe Hur, and Sungroh Yoon.
\newblock Joint contrastive learning for unsupervised domain adaptation, 2020.

\bibitem{DBLP:journals/corr/abs-1912-01703}
Adam Paszke, Sam Gross, Francisco Massa, Adam Lerer, James Bradbury, Gregory
  Chanan, Trevor Killeen, Zeming Lin, Natalia Gimelshein, Luca Antiga, Alban
  Desmaison, Andreas K{\"{o}}pf, Edward Yang, Zach DeVito, Martin Raison,
  Alykhan Tejani, Sasank Chilamkurthy, Benoit Steiner, Lu Fang, Junjie Bai, and
  Soumith Chintala.
\newblock Pytorch: An imperative style, high-performance deep learning library.
\newblock {\em CoRR}, abs/1912.01703, 2019.

\bibitem{Richter_2016_ECCV}
Stephan~R. Richter, Vibhav Vineet, Stefan Roth, and Vladlen Koltun.
\newblock Playing for data: {G}round truth from computer games.
\newblock In Bastian Leibe, Jiri Matas, Nicu Sebe, and Max Welling, editors,
  {\em European Conference on Computer Vision (ECCV)}, volume 9906 of {\em
  LNCS}, pages 102--118. Springer International Publishing, 2016.

\bibitem{7780721}
G. {Ros}, L. {Sellart}, J. {Materzynska}, D. {Vazquez}, and A.~M. {Lopez}.
\newblock The synthia dataset: A large collection of synthetic images for
  semantic segmentation of urban scenes.
\newblock In {\em 2016 IEEE Conference on Computer Vision and Pattern
  Recognition (CVPR)}, pages 3234--3243, 2016.

\bibitem{DBLP:conf/cvpr/SaitoWUH18}
Kuniaki Saito, Kohei Watanabe, Yoshitaka Ushiku, and Tatsuya Harada.
\newblock Maximum classifier discrepancy for unsupervised domain adaptation.
\newblock In {\em 2018 {IEEE} Conference on Computer Vision and Pattern
  Recognition, {CVPR} 2018, Salt Lake City, UT, USA, June 18-22, 2018}, pages
  3723--3732. {IEEE} Computer Society, 2018.

\bibitem{10.5555/3157096.3157333}
Ozan Sener, Hyun~Oh Song, Ashutosh Saxena, and Silvio Savarese.
\newblock Learning transferrable representations for unsupervised domain
  adaptation.
\newblock In {\em Proceedings of the 30th International Conference on Neural
  Information Processing Systems}, NIPS'16, page 2118–2126, Red Hook, NY,
  USA, 2016. Curran Associates Inc.

\bibitem{DBLP:conf/iclr/ShuBNE18}
Rui Shu, Hung~H. Bui, Hirokazu Narui, and Stefano Ermon.
\newblock A {DIRT-T} approach to unsupervised domain adaptation.
\newblock In {\em 6th International Conference on Learning Representations,
  {ICLR} 2018, Vancouver, BC, Canada, April 30 - May 3, 2018, Conference Track
  Proceedings}. OpenReview.net, 2018.

\bibitem{su2020gradient}
Peng Su, Shixiang Tang, Peng Gao, Di Qiu, Ni Zhao, and Xiaogang Wang.
\newblock Gradient regularized contrastive learning for continual domain
  adaptation, 2020.

\bibitem{DBLP:conf/cvpr/TsaiHSS0C18}
Yi{-}Hsuan Tsai, Wei{-}Chih Hung, Samuel Schulter, Kihyuk Sohn, Ming{-}Hsuan
  Yang, and Manmohan Chandraker.
\newblock Learning to adapt structured output space for semantic segmentation.
\newblock In {\em 2018 {IEEE} Conference on Computer Vision and Pattern
  Recognition, {CVPR} 2018, Salt Lake City, UT, USA, June 18-22, 2018}, pages
  7472--7481. {IEEE} Computer Society, 2018.

\bibitem{DBLP:journals/corr/abs-1807-03748}
A{\"{a}}ron van~den Oord, Yazhe Li, and Oriol Vinyals.
\newblock Representation learning with contrastive predictive coding.
\newblock {\em CoRR}, abs/1807.03748, 2018.

\bibitem{Vu_2019_CVPR}
Tuan-Hung Vu, Himalaya Jain, Maxime Bucher, Matthieu Cord, and Patrick Perez.
\newblock Advent: Adversarial entropy minimization for domain adaptation in
  semantic segmentation.
\newblock In {\em Proceedings of the IEEE/CVF Conference on Computer Vision and
  Pattern Recognition (CVPR)}, June 2019.

\bibitem{DBLP:journals/corr/abs-1804-05827}
Zuxuan Wu, Xintong Han, Yen{-}Liang Lin, Mustafa~G{\"{o}}khan Uzunbas, Tom
  Goldstein, Ser{-}Nam Lim, and Larry~S. Davis.
\newblock {DCAN:} dual channel-wise alignment networks for unsupervised scene
  adaptation.
\newblock {\em CoRR}, abs/1804.05827, 2018.

\bibitem{DBLP:conf/iccv/YueZZSKG19}
Xiangyu Yue, Yang Zhang, Sicheng Zhao, Alberto~L. Sangiovanni{-}Vincentelli,
  Kurt Keutzer, and Boqing Gong.
\newblock Domain randomization and pyramid consistency: Simulation-to-real
  generalization without accessing target domain data.
\newblock In {\em 2019 {IEEE/CVF} International Conference on Computer Vision,
  {ICCV} 2019, Seoul, Korea (South), October 27 - November 2, 2019}, pages
  2100--2110. {IEEE}, 2019.

\bibitem{DBLP:journals/corr/abs-1910-13049}
Qiming Zhang, Jing Zhang, Wei Liu, and Dacheng Tao.
\newblock Category anchor-guided unsupervised domain adaptation for semantic
  segmentation.
\newblock {\em CoRR}, abs/1910.13049, 2019.

\bibitem{DBLP:journals/corr/ZhaoSQWJ16}
Hengshuang Zhao, Jianping Shi, Xiaojuan Qi, Xiaogang Wang, and Jiaya Jia.
\newblock Pyramid scene parsing network.
\newblock {\em CoRR}, abs/1612.01105, 2016.

\bibitem{DBLP:journals/corr/ZhuPIE17}
Jun{-}Yan Zhu, Taesung Park, Phillip Isola, and Alexei~A. Efros.
\newblock Unpaired image-to-image translation using cycle-consistent
  adversarial networks.
\newblock {\em CoRR}, abs/1703.10593, 2017.

\bibitem{CycleGAN2017}
Jun-Yan Zhu, Taesung Park, Phillip Isola, and Alexei~A Efros.
\newblock Unpaired image-to-image translation using cycle-consistent
  adversarial networks.
\newblock In {\em Computer Vision (ICCV), 2017 IEEE International Conference
  on}, 2017.

\end{thebibliography}
}

\end{document}